\documentclass{article}

\IfFileExists{neurips_2025.sty}{%
    \usepackage[preprint]{neurips_2025}%
}{%
    \PackageWarningNoLine{main}{neurips_2025.sty not found; using plain article fallback.}%
    \usepackage[margin=1in]{geometry}%
    \usepackage{titling}%
    \setlength{\droptitle}{-2em}%
}

\usepackage[utf8]{inputenc}
\usepackage[T1]{fontenc}
\usepackage{microtype}
\usepackage{hyperref}
\usepackage{url}
\usepackage{booktabs}
\usepackage{amsmath, amssymb}
\usepackage{graphicx}
\usepackage{subcaption}
\usepackage{xcolor}
\usepackage{algorithm}
\usepackage{algpseudocode}
\usepackage{multirow}
\usepackage{enumitem}
\usepackage{authblk}

\hypersetup{colorlinks=true, linkcolor=blue!60!black, citecolor=blue!60!black, urlcolor=teal}

\newcommand{\mlptok}{\textsc{MLP-token}}
\newcommand{\lintok}{\textsc{Linear-token}}
\newcommand{\dualhead}{\textsc{Dual-head}}
\newcommand{\auc}{\textsc{AUC}}
\newcommand{\fone}{\ensuremath{F_1}}
\newcommand{\autores}{\textsc{AutoResearch}}

\title{Stop Early, Spend Less: Hidden-State Probes as a \\
       Practical Recipe for Streaming Moderation of LLM Outputs}

\author[1]{Huizhen Shu}
\author[2]{Xuying Li}
\author[2]{Piao Xue}

\affil[1]{ModelOneAI,{shuhuizhen@modeloneai.cn}}
\affil[2]{yunshanai, \texttt{\{lixuying,piaoxue\}@yunshanai.org.cn}}

\begin{document}
\maketitle

\begin{abstract}
Deploying large language models (LLMs) in user-facing products
requires \emph{output} safety filtering that can act \emph{during}
generation. The prevailing recipe chains a second moderation
model---a moderation API, Llama~Guard, or programmable guardrails---after
the generator, but such a guard reads the \emph{completed} text: it
roughly doubles serving cost and can only flag a violation after the
unsafe content has reached the user. We start from a simple
observation: the signal needed for moderation already lives inside the
generating model. We train lightweight \emph{token-level probes} on the
model's own hidden states and aggregate their per-token unsafe scores
into either an offline verdict on a finished response or an online
decision that fires the moment a generation turns unsafe. Because the
probe reuses activations the model has already computed, it needs no
second forward pass and runs inside the vLLM decode loop.
A small probe on a single mid-network layer recovers the large majority
of a strong guard model's verdicts; it is best understood as a cheap,
inlined surrogate for that guard rather than a more accurate moderator,
so its value lies in cost and timing rather than raw accuracy.
Streaming, it halts or rewrites an unsafe response before most of it
reaches the user, replacing a roughly half-second, end-of-response
guard call with a sub-millisecond, per-token check. Benchmarked against
both a post-hoc and a streaming guard model on neutral human labels,
the probe is the weaker classifier but adds two-to-four orders of
magnitude less moderation compute and near-zero latency. We distill our
sweeps into a concrete deployment recipe---which layer to read, how to
aggregate scores, how often to probe, and how to set the trigger---so
the method drops into an existing serving stack. Finally, since the
probe's linear component is a direction in residual space, the same
hidden state could in principle be written to for activation steering;
we sketch this read/write symmetry as a preliminary direction for
future work.
\end{abstract}

\section{Introduction}
\label{sec:intro}

LLM providers must intercept unsafe model outputs before they reach
end users. The standard recipe is to chain a separate classifier
after generation: a moderation API, a guard model such as
Llama~Guard~\citep{inan2023llamaguard}, or rule-based
guardrails~\citep{rebedea2023nemo}. Two limitations motivate this work.
\textbf{(i) Latency.} A second model must read the full response
before deciding, so the user has already seen the unsafe content by
the time the verdict arrives. In our measurements, a
\texttt{Qwen3Guard-Gen-8B}~\citep{zhao2025qwen3guardtechnicalreport} post-hoc moderator adds roughly $480$~ms
per response. \textbf{(ii) Cost.} Running an extra LLM as a moderator
can double serving cost, which becomes prohibitive for long, streamed
responses.

Prior work on representation engineering~\citep{zou2023repe},
truthfulness probes~\citep{li2023inferencetime}, latent-knowledge
classifiers~\citep{burns2023discovering}, and sleeper-agent
detection~\citep{macdiarmid2024sleeper} suggests that simple
classifiers on the hidden states of a base LLM can recover surprisingly
strong signals about model behaviour. Most directly, Anthropic's
Constitutional Classifiers~\citep{sharma2025constitutional} and their
successor Constitutional~Classifiers++~\citep{cunningham2026ccpp}
deploy lightweight activation probes as a cheap first-stage screen
inside a production jailbreak-defence cascade, reusing the model's own
internal states almost for free. Our setting pushes this idea in two
practically important directions. First, we run the probe per
token, during generation, so it can stop or rewrite a response before
it finishes rather than only screen a completed exchange. Second, we
let the probe be the moderation decision instead of a
first-stage filter that escalates flagged exchanges to a heavier
classifier; this keeps the second model off the inference path and
turns probe cost into an explicit, tunable operating point. We build on
this use of internal probes and take the line of work to its
operational conclusion:
\emph{the same hidden states that already exist
inside the serving model can be probed cheaply, per token, to decide
whether the current generation is becoming unsafe.}

\paragraph{Contributions.}
\begin{enumerate}[leftmargin=*]
\item \textbf{A self-monitoring formulation.} We cast output safety as
      a probing problem on the generator's own hidden states,
      supervised by an existing guard model's labels, with a per-token
      weighted-BCE objective and a \textsc{topK} pool that concentrates
      supervision on the few tokens where unsafe content actually
      surfaces (Sec.~\ref{sec:method}). The resulting probe distills the
      guard into the generator's forward pass and needs no second model
      at inference.
\item \textbf{A streaming safety monitor, as the main contribution.}
      We integrate the probe into the vLLM decode loop so that it emits
      a verdict per token and can stop or rewrite a response before the
      user sees it. We evaluate it end to end on three benchmarks and
      characterise its behaviour across probe-interval, score-aggregation,
      and trigger modes (Sec.~\ref{sec:online}), distilling a practical
      deployment recipe and a tunable accuracy--cost trade-off. We also
      benchmark its cost and latency head-to-head against post-hoc and
      streaming guard models on neutral human labels
      (Sec.~\ref{sec:online_cost}).
\item \textbf{An autonomous architecture search.} An
      \autores-style agent searches the probe design space on top of a
      hand-designed baseline and converges on a \dualhead\ probe, which
      we then retrain at full data scale
      (Secs.~\ref{sec:train_hand}--\ref{sec:auto});
      we report offline results, including a per-category breakdown,
      in Sec.~\ref{sec:offline}.
\end{enumerate}
Beyond these contributions, we note a conceptual bridge to
\emph{activation steering}: because the linear branch of \dualhead\ is a
direction in residual space, the hidden state read for detection can in
principle be written to in order to re-route the generator. We report a
preliminary prototype but leave a quantitative evaluation to future
work (Sec.~\ref{sec:discuss}).

\section{Related work}
\label{sec:related}
\textbf{Probing classifiers.}
Linear and shallow MLP probes have a long history as diagnostic tools
for transformer representations \citep{alain2017understanding,
tenney2019bert}. Recent work uses them operationally: contrast-consistent
search~\citep{burns2023discovering} extracts latent truthfulness;
\citet{li2023inferencetime} and~\citet{zou2023repe} use probe directions
for steering; \citet{macdiarmid2024sleeper} demonstrate that very simple
probes catch backdoored sleeper-agent models.
Closest to our setting, Constitutional~Classifiers++~\citep{cunningham2026ccpp}
use a linear activation probe as a cheap first-stage screen that reuses
the model's internal states, escalating only flagged exchanges to a
heavier classifier; our streaming monitor pushes the same idea to a
per-token, intervene-during-generation regime.

\textbf{Safety filtering for LLMs.}
Production stacks rely on a second classifier:
Llama~Guard~\citep{inan2023llamaguard} fine-tunes a separate LLM as a
moderator; OpenAI's moderation pipeline~\citep{markov2023holistic}
uses a calibrated text classifier; NeMo~Guardrails~\citep{rebedea2023nemo}
adds programmable rails; and Constitutional
Classifiers~\citep{sharma2025constitutional} train input/output
classifiers on constitution-guided synthetic data. All of these read
the completed text and therefore cannot intervene during
streaming.

\textbf{Safety benchmarks.}
BeaverTails~\citep{ji2023beavertails} and
ToxicChat~\citep{lin2023toxicchat} provide English content-safety
labels; we additionally use a Chinese, fine-grained 49-category
compliance dataset (\textsc{Zhuyi}) that exposes both broad harm types
(state security, discrimination, infringement) and adversarial
categories (\textsc{leetspeak\_homoglyph}, \textsc{pinyin\_homophone},
\textsc{roleplay\_jailbreak}, \textsc{injection\_and\_hijack}, etc.).

\textbf{Autonomous research loops.}
Our autonomous-search workflow follows the spirit of
\textsc{AutoResearch}~\citep{karpathy2026autoresearch}: an LLM agent
holds the only writable file in the repository, edits it, runs
training, reads the metric, and decides whether to keep or revert
the change.

\section{Method}
\label{sec:method}

\subsection{Background and notation}
Let $\mathcal{M}$ be a frozen autoregressive LLM with $L$ transformer
layers. For an input sequence $x_{1{:}T}$ we denote the hidden state
of layer $\ell$ at position $t$ as $h^{\ell}_t \in \mathbb{R}^{H}$, and
the attention mask as $m_t \in \{0,1\}$. We aim to produce a per-token
\emph{unsafe logit} $z_t \in \mathbb{R}$ such that
$\sigma(z_t) \approx \Pr[\text{token~$t$ is part of an unsafe
generation}]$, and a sequence-level decision
$\hat{y} = \mathbf{1}\{\text{Agg}(z_{1:T}) \ge 0\}$.

\subsection{Probe heads}
\label{sec:heads}
We train three token-level heads that share the LLM forward pass and
extract a single layer's hidden state $h^{\ell}_t$:
\begin{align*}
\textbf{\lintok:}\quad & z_t = w^\top h^{\ell}_t + b,
   \quad w \in \mathbb{R}^{H} \\
\textbf{\mlptok:}\quad & z_t = w_2^\top \,\sigma\!\left(W_1 h^{\ell}_t + b_1\right) + b_2,
   \quad W_1 \in \mathbb{R}^{d \times H}\\
\textbf{\dualhead:}\quad & z_t = \alpha \cdot z_t^{\text{mlp}} + (1-\alpha) \cdot z_t^{\text{lin}}
\end{align*}
with $d=1024$ throughout and $\sigma$ being \textsc{ReLU} after the
autonomous search (\textsc{GELU} in the hand-designed baseline). The
\dualhead\ head is a linear convex combination of a two-layer MLP and
a single linear branch, both reading the same $h^{\ell}_t$. The
LLM parameters are frozen ($\nabla_{\theta_{\mathcal{M}}}=0$); only
the heads are trained.

\subsection{Per-token weighted BCE loss}
\label{sec:loss}
Following the use of lightweight activation probes for safety
screening in Constitutional~Classifiers++~\citep{cunningham2026ccpp},
we use $z_t$ itself to softly select the most informative tokens of
each sequence. Let
\textsc{Pool}$_K$ denote either a length-$M$ sliding-window mean
(\textsc{SWiM}$_M$) or a \textsc{topK} pooling that retains the $K$
largest per-token logits. We use both: \textsc{SWiM} during the hand-
designed exploration and \textsc{topK} once the autonomous search
identifies it (Sec.~\ref{sec:auto}). Writing the pooled logit as
$\bar{z}_t$,
\begin{equation*}
w_t = \frac{\exp(\bar{z}_t / \tau)}
            {\sum_{t'\in V} \exp(\bar{z}_{t'} / \tau)}
\,, \quad
\mathcal{L} = \sum_{t \in V} w_t \cdot
              \mathrm{BCE}(\bar{z}_t,\, y)
\end{equation*}
where $V=\{t: m_t = 1\}$ is the set of non-pad positions, $y\in\{0,1\}$
is the \emph{sequence}-level label (Safe$=0$, Unsafe$=1$), and
$\tau=1$ unless stated otherwise.

\subsection{Sequence-level aggregation at inference}
\label{sec:agg}
Given $\{z_t\}_{t\in V}$ we form a single sequence score via one of
\textsc{max}, \textsc{topk\_mean} ($k{=}3$),
\textsc{window\_max} (last $w{=}20$ tokens),
\textsc{percentile} ($p{=}90$),
\textsc{logsumexp} (temperature $\tau{=}1$), or
\textsc{softmax} (weighted mean with $w_t \propto \exp(z_t/\tau)$).
\textsc{Max} is the natural choice for online monitoring because it
matches a streaming early-stop rule ``trigger the first time
$z_t \ge 0$''; the others are studied for offline robustness.

\subsection{Online streaming monitor}
\label{sec:streaming}
For online monitoring the probe shares the forward pass of the
generator: at each decoding step the cached hidden state of layer
$\ell$ is forwarded through the (small) probe head, giving $z_t$ in
$\mathcal{O}(H d)$ time per token. We register a \texttt{forward\_hook}
on the chosen layer inside vLLM~\citep{kwon2023vllm}, and let the
worker maintain a small running state $s_t = \phi(z_{\le t})$ for one
of seven \emph{score-aggregation} modes
(\textsc{raw}, \textsc{ema}, \textsc{softmax},
\textsc{window\_max}, \textsc{topk\_mean},
\textsc{percentile}, \textsc{logsumexp}).
A separate \emph{trigger mode} then converts $s_t$ to a binary firing
decision, ranging from firing on a single threshold crossing
(\textsc{instant}) to requiring $K$ cumulative or consecutive
crossings (\textsc{cumulative}, \textsc{consecutive}), a leaky counter
(\textsc{decay}), or an accumulated-excess budget (\textsc{budget}).
A separate \emph{warmup} phase (steps $\le W_0$) optionally uses a
higher threshold to suppress false positives from chain-of-thought
prefixes. To reduce per-token overhead we additionally allow a
\emph{probe interval} \texttt{pi}: the hook only runs every \texttt{pi}
decode steps. The cost--accuracy trade-off of \texttt{pi} is the focus
of Sec.~\ref{sec:online}.

\section{Experimental setup}
\label{sec:setup}

\paragraph{Backbones.}
All reported results use \texttt{DeepSeek-R1-0528-Qwen3-8B}, a
DeepSeek-R1~\citep{deepseekr1} reasoning model distilled into a
Qwen3~\citep{qwen3} backbone. We also trained probes on other
backbones, including \texttt{Qwen3-8B} and
\texttt{Qwen3-30B-A3B-Thinking-2507}~\citep{qwen3}, and observed
qualitatively similar behaviour---a sharp mid-network rise in probe
quality and the same head and aggregation preferences; for brevity we
omit these runs and report the \texttt{DeepSeek-R1-0528-Qwen3-8B}
results throughout. The backbone is loaded with \texttt{bfloat16} via
\texttt{transformers.AutoModelForCausalLM} and frozen during probe
training.

\paragraph{Datasets.}
\textsc{Zhuyi} is an in-house Chinese safety/compliance dataset with
49 fine-grained categories grouped under broader headings (state
security, discrimination, infringement, rights, jailbreak,
obfuscation, etc.). Each example consists of a prompt and a response
obtained by querying an LLM with that prompt; the sequence-level risk
label in \{Safe, Unsafe\} is assigned by \texttt{Qwen3Guard-Gen-8B}, and
is accompanied by per-category soft scores in $[0,1]$. We partition the
data into disjoint training, validation ($5{,}000$ examples), and
held-out test ($2{,}444$ examples) splits, and report all offline
numbers on the test split.
For online evaluation we additionally use
\textsc{BeaverTails}~\citep{ji2023beavertails} and
\textsc{CSSBench}~\citep{zhou2026cssbenchevaluatingsafetylightweight},
a benchmark of Chinese-specific adversarial safety patterns.

\paragraph{Training.}
We sample $10^5$ training rows per epoch from the training split; the
sampler is label-balanced and supplements the main pool with a
length-filtered short-text pool (prompt/response $<100$ characters)
so that the probe sees ample short adversarial prompts. We train for
3--4 epochs with AdamW, learning rate $1\!\times\!10^{-4}$,
$\text{batch}{=}8$--$64$, warmup ratio $0.05$--$0.1$, weight decay
$0.01$, gradient clipping at $1.0$, \texttt{bfloat16}, max-length
$1024$. Evaluation and checkpointing happen every $4000$ steps; the
best model by validation \fone\ is restored at the end of training.
The autonomous search of Sec.~\ref{sec:auto} works on a
cached-hidden-states fast path: a data-preparation step pre-extracts
layer 20/21/22 activations for $10{,}000$ training and $2{,}000$
validation rows so that each agent iteration takes 1--3 minutes
instead of $\sim 7$ minutes.

\paragraph{Metrics.}
Offline: accuracy, binary \fone, precision, recall, and \auc\ on
the \textsc{Zhuyi} test split.
Online: \fone, precision, recall, plus operational signals
(average stop step, average tokens saved, probe-call savings).

\section{Training, stage 1: hand-designed probe ablation}
\label{sec:train_hand}

\subsection{Layer selection}
\label{sec:layer}
Sweeping the \mlptok\ probe (max aggregation) across layers on the
held-out test split (top block of Table~\ref{tab:offline}), all metrics
rise sharply between layers~18 and~21 and plateau around layer~21
(\fone~$=0.941$, \auc~$=0.942$), then saturate beyond layer~24: the
relevant ``unsafe'' feature is already linearly readable by
mid-network. We adopt layer~21 as the default \emph{primary} layer for
online inference; the autonomous search of Sec.~\ref{sec:auto} also
converges on layer~21.

\subsection{Linear vs.\ MLP head}
\label{sec:lin_vs_mlp}
Replacing the \mlptok\ head with a single \lintok\ projection drops
\fone\ from $0.941$ to $0.892$ at layer~21 (Table~\ref{tab:offline},
matched training budget), with the gap widening on fine-grained
obfuscation categories where the unsafe signal is least linearly
separable. The linear head nonetheless retains a property we exploit in
Sec.~\ref{sec:discuss}: its weight vector $w$ \emph{is} a usable
steering direction in residual space.

\subsection{Token-logit aggregation}
\begin{figure}[t]
\centering
\includegraphics[width=0.78\linewidth]{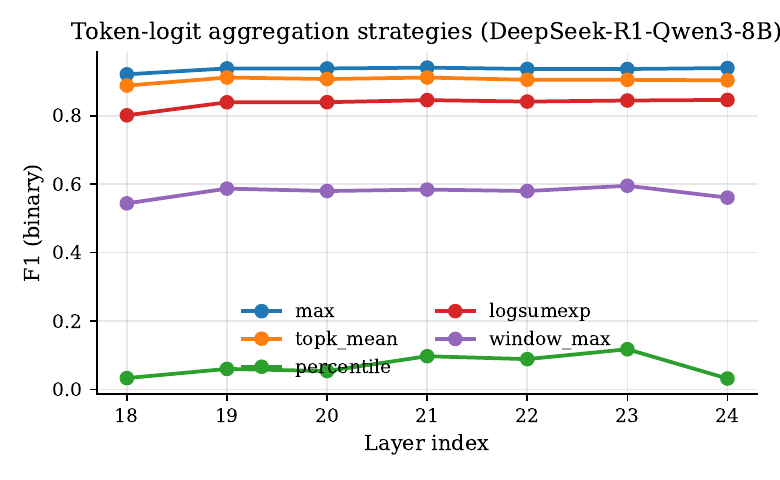}
\caption{Token-logit aggregation strategies vs.\ layer
(\mlptok, test set). \textsc{Max} matches online behaviour;
\textsc{topk\_mean} and \textsc{window\_max} are the most robust
offline aggregations.}
\label{fig:token_agg}
\end{figure}

We hold the probe fixed (\mlptok, layer~21) and vary the inference-time
aggregation in Fig.~\ref{fig:token_agg}.
\textsc{Max} dominates: it is the natural ``trigger as soon as any
token looks unsafe'' rule and matches an online early-stop policy.
\textsc{Topk\_mean} and \textsc{window\_max} closely track \textsc{max}
($\Delta$\fone$<0.005$) and are more robust to noisy single-token
spikes. \textsc{Percentile} ($p{=}90$) collapses for long sequences
because the 90th percentile of safe prefixes can dominate the
unsafe suffix.
\textsc{Logsumexp} is between \textsc{topk\_mean} and \textsc{max}.

\section{Training, stage 2: \autores\ search over probe architecture}
\label{sec:auto}

After the hand-designed ablation of Sec.~\ref{sec:train_hand}
converged on a strong but conservative configuration (\mlptok\ layer~21,
\textsc{max} aggregation, sliding-window pooling), we let an LLM
agent drive the next round of search autonomously, in the style of
\citet{karpathy2026autoresearch}. The setup mirrors that work
closely:

\begin{itemize}[leftmargin=*]
\item A fixed data-preparation step pre-extracts hidden states for
      layers $\{20, 21, 22\}$ on $10{,}000$ training and $2{,}000$
      validation rows into a local cache.
\item A single training module---containing the probe head, optimiser,
      loss, and training loop---is the only file the agent may edit.
\item A protocol specification fixes the experimental rules (one
      hypothesis per iteration, keep iff the validation \fone\ improves,
      revert otherwise, never stop the loop).
\end{itemize}

We ran a single overnight loop of $177$ training trials. The
agent accepted $11$ of them (validation \fone\ strictly improved) and
rejected $163$ (\fone\ matched or regressed); $3$
runs crashed and were discarded. Table~\ref{tab:auto_keep} reports the
accepted chain. The whole search lifted validation \fone\ from
$0.9264$ (the hand-designed baseline) to $0.9495$ ($+0.0231$,
$\sim 25$\% relative error reduction on $1-\text{\fone}$).

\begin{table}[t]
\centering
\small
\caption{The accepted changes along the \autores\ loop, in
chronological order. Step~0 is the hand-designed baseline; each
subsequent step is the smallest single-variable change that improved
validation \fone\ on the cached validation subset (N=2000). The final
step is the global best.}
\label{tab:auto_keep}
\begin{tabular}{lcccl}
\toprule
Step & val \fone\ & val \auc\ & $\Delta$ vs.\ start & Change \\
\midrule
0 (baseline) & 0.9264 & 0.9808 & --- & \mlptok, MLP\_HIDDEN=1024, 2 layers, LR=$10^{-4}$, 3 epochs \\
1 & 0.9281 & 0.9809 & +0.0017 & DROPOUT $0\to0.1$ \\
2 & 0.9301 & 0.9804 & +0.0037 & EPOCHS $3\to4$ \\
3 & 0.9310 & 0.9805 & +0.0046 & activation $\to$ ReLU \\
4 & 0.9386 & 0.9874 & +0.0122 & aggregation $\to$ \textsc{topk16} \\
5 & 0.9401 & 0.9872 & +0.0137 & WARMUP\_RATIO $\to 0.1$ \\
6 & 0.9403 & 0.9876 & +0.0139 & DROPOUT $\to 0.05$ \\
7 & 0.9415 & 0.9870 & +0.0151 & DROPOUT $\to 0.15$ \\
8 & 0.9440 & 0.9873 & +0.0176 & DROPOUT $\to 0.18$ \\
9 & 0.9476 & 0.9886 & +0.0212 & \dualhead\ architecture, $\alpha=0.7$ \\
\textbf{10} & \textbf{0.9495} & \textbf{0.9886} & \textbf{+0.0231} & \dualhead\ $\alpha=0.8$ \\
\bottomrule
\end{tabular}
\end{table}

\paragraph{What the agent found.}
Two changes dominate the gain:
\begin{enumerate}[leftmargin=*]
\item \textbf{\textsc{topk16} loss-time pooling} (step~4,
$+0.0076$). Instead of pooling the per-token logits
with a length-$M$ sliding-window mean, the agent kept only the
$K{=}16$ tokens with the largest $z_t$ per sequence before computing
the weighted-BCE loss of Sec.~\ref{sec:loss}. The intuition is the
mirror of \textsc{max}-at-inference: the unsafe signal lives in a
small fraction of the response, and asking the head to fit the
average over all tokens dilutes it. The agent also explicitly
verified $K \in \{8, 12, 24, 32\}$; all of them are worse than
$K{=}16$ but by $\le 0.002$ \fone, indicating $K$ is robust within
that range.
\item \textbf{\dualhead\ architecture} (steps~9 and~10,
$+0.0089$). The final architecture interpolates the
MLP branch and a single linear branch on the \emph{same} hidden
state: $z_t = \alpha z_t^{\text{mlp}} + (1-\alpha) z_t^{\text{lin}}$.
The agent swept $\alpha$ in $\{0.1, 0.3, 0.5, 0.7, 0.8, 0.85, 0.9\}$
and identified $\alpha = 0.8$ as the optimum. We hypothesise that
the linear branch behaves as a strong prior in the early steps of
training (the linear direction is recovered very quickly, see
Sec.~\ref{sec:lin_vs_mlp}), while the MLP branch is allowed to
specialise on the remaining non-linear cases.
\end{enumerate}
The remaining keeps are mild regulariser adjustments
(\textsc{Dropout} $\in [0.15, 0.18]$, \textsc{ReLU}, warmup ratio
$0.1$). Several other directions explored by the agent
\emph{did not} improve validation \fone\ and were reverted; we summarise
the negative results below because they are themselves informative.

\paragraph{What the agent ruled out.}
Each of the following was tested in $\ge 3$ trials and rejected:
larger MLP hidden ($1{,}536$ / $2{,}048$), three-layer MLP,
\texttt{LayerNorm}, focal loss, label smoothing, attention pooling
heads (\textsc{attn}, \textsc{attn\_mlp}), residual MLP blocks
(\textsc{res\_token}), gated MLP blocks (\textsc{gated\_token}),
dual-path aggregation, batch sizes outside $\{32, 64\}$, learning
rates outside $\{8\!\cdot\!10^{-5}, 10^{-4}\}$, and seed swaps. We
take this as evidence that \dualhead\ is not merely the product of
exhaustive hyper-parameter tuning: the agent did try many things, and
only the two architectural changes above survived. The final
configuration is \dualhead\ at layer~21 (MLP hidden $1024$, two layers,
\textsc{ReLU}, dropout $0.18$, \textsc{topk16} training-time pool,
$\alpha{=}0.8$), trained with weighted-BCE for $4$ epochs.

\begin{figure}[t]
\centering
\includegraphics[width=0.95\linewidth]{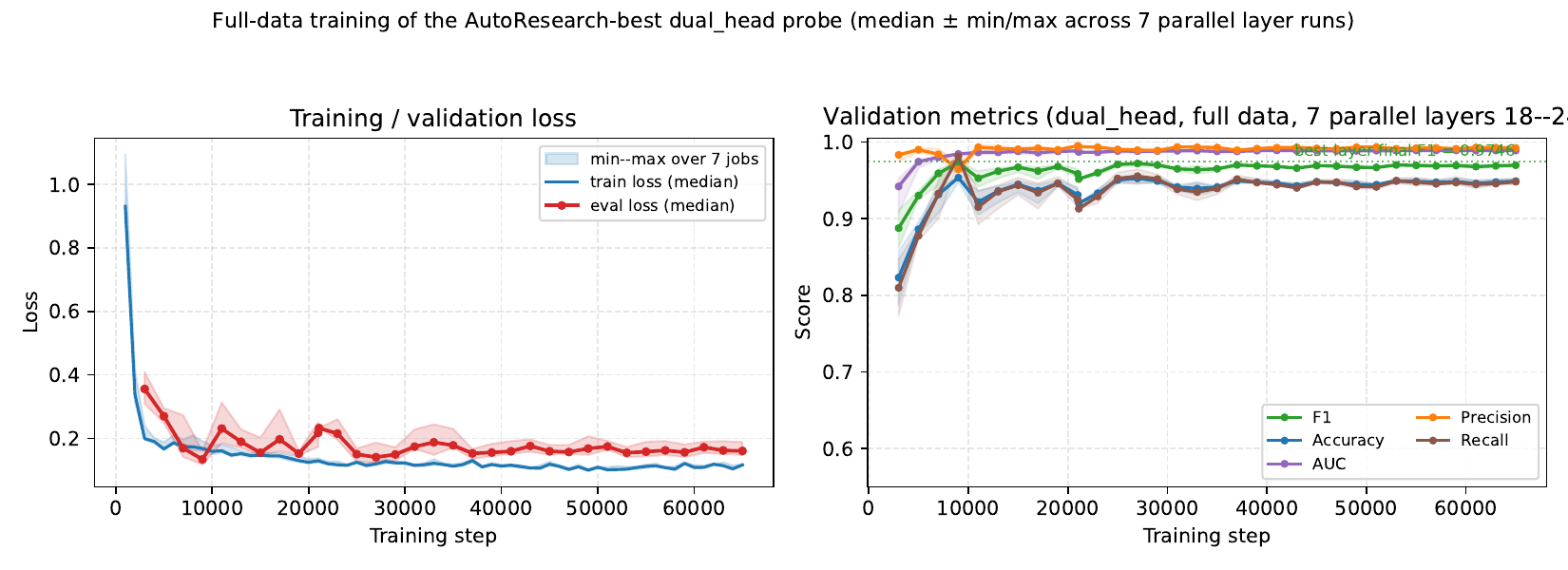}
\caption{Full-data training of the AutoResearch-best \dualhead\ probe.
Left: training loss (median $\pm$ min/max over seven parallel layer
jobs $\ell \in \{18,\ldots,24\}$) and held-out validation loss at the
same eval steps. Right: held-out validation \fone\ / accuracy / AUC
/ precision / recall, again median $\pm$ min/max across the seven
parallel layers. The horizontal dotted line marks the best layer's
final \fone~$=0.9746$.}
\label{fig:training_curve}
\end{figure}

\paragraph{Full-data retraining.}
The autonomous search of Table~\ref{tab:auto_keep} was carried out on
the cached $10{,}000$-row training subset for fast iteration. Once
the architecture (\dualhead, $\alpha{=}0.8$) and the hyper-parameter
recipe had stabilised we re-trained the probe on the \emph{full}
\textsc{Zhuyi} training split ($\sim 600$K rows, $72{,}423$ optimiser
steps over $3$ epochs), still freezing the LLM. To obtain a
layer-sensitivity sweep at no additional cost, we launched seven jobs
in parallel, one per layer index in $\{18, 19, \ldots, 24\}$, sharing
the same \dualhead\ recipe. Figure~\ref{fig:training_curve} plots
training loss and the five validation metrics, taking the median
$\pm\,[\min, \max]$ over the seven parallel layers at each
eval-step. All seven jobs converge above \fone~$=0.97$ within the
first $\sim 30$K steps and continue improving slowly until the end of
epoch~3; the best layer reaches a final \fone\ of $0.9746$
(\auc~$=0.9865$, precision~$=0.985$, recall~$=0.965$) on the
validation split, $\sim 3$ points above the
hand-designed \mlptok\ baseline of Sec.~\ref{sec:train_hand}.

\section{Offline evaluation}
\label{sec:offline}

\paragraph{Setting.}
The offline evaluation feeds the probe a complete
(\texttt{prompt}, \texttt{response}) pair, runs \emph{one} forward
pass through the frozen LLM, and aggregates per-token unsafe logits
into a sequence verdict. This mirrors how a post-hoc moderator
(Llama~Guard) is normally used, except that the moderator and the
generator share parameters. Since the probe is trained on the guard
model's labels, the \fone\ reported here measures \emph{agreement with
the guard} rather than ground-truth accuracy; the probe is a distilled
surrogate and is not expected to exceed the guard it learns from.

\paragraph{Main results.}
Table~\ref{tab:offline} reports the offline numbers. The
hand-designed \mlptok\ at layer~21 with \textsc{max} aggregation
already agrees closely with the guard on the held-out test set
($N{=}2444$): \fone~$=0.941$, \auc~$=0.942$. The autonomous-search \dualhead\
configuration (Sec.~\ref{sec:auto}) further improves \fone\ on the
cached validation subset to $0.9495$; re-training the same \dualhead\ recipe
on the full training split and evaluating on the
validation split raises it again to
\fone~$=0.9746$ (\auc~$=0.987$) for the best layer
(Fig.~\ref{fig:training_curve}). The hand-designed \mlptok\ remains
the configuration we deploy in the online experiments of
Sec.~\ref{sec:online}: the streaming worker was integrated against
it, and the full-data \dualhead\ checkpoint has not yet been re-run
through vLLM.

\begin{table}[t]
\centering
\small
\caption{Offline probe results on \textsc{Zhuyi}. The top block is
the hand-designed sweep evaluated on the held-out test split.
The bottom block is the autonomous-search
\dualhead\ configuration: in its cached form
(Sec.~\ref{sec:auto}, validation subset) and after retraining on the
full training split (Sec.~\ref{sec:auto}, validation split).}
\label{tab:offline}
\begin{tabular}{lcccccc}
\toprule
Probe head & Layer & Agg.\ & ACC & \fone\ & Precision & \auc \\
\midrule
\multicolumn{7}{l}{\textit{Hand-designed (Sec.~\ref{sec:train_hand}), test split}} \\
\mlptok & 18 & max          & 0.924 & 0.921 & 0.950 & 0.924 \\
\mlptok & 19 & max          & 0.940 & 0.939 & 0.956 & 0.940 \\
\mlptok & 20 & max          & 0.940 & 0.938 & 0.959 & 0.940 \\
\textbf{\mlptok} & \textbf{21} & \textbf{max} & \textbf{0.942} & \textbf{0.941} & \textbf{0.961} & \textbf{0.942} \\
\mlptok & 22 & max          & 0.938 & 0.937 & 0.952 & 0.938 \\
\mlptok & 24 & max          & 0.941 & 0.940 & 0.952 & 0.941 \\
\lintok & 21 & max          & 0.897 & 0.892 & 0.932 & 0.897 \\
\mlptok & 21 & topk\_mean   & 0.918 & 0.912 & 0.984 & 0.918 \\
\mlptok & 21 & window\_max  & 0.867 & 0.846 & 0.995 & 0.866 \\
\mlptok & 21 & logsumexp    & 0.706 & 0.584 & 0.987 & 0.705 \\
\mlptok & 21 & percentile   & 0.528 & 0.097 & 1.000 & 0.525 \\
\midrule
\multicolumn{7}{l}{\textit{Autonomous search (Sec.~\ref{sec:auto})}} \\
\dualhead, $\alpha{=}0.8$ & 21 & topk16-loss + max-eval &
   --- & 0.9495 & --- & 0.9886 \\
\textbf{\dualhead, $\alpha{=}0.8$ (full data)} & \textbf{18--24 best}
   & \textbf{topk16-loss + max-eval}
   & --- & \textbf{0.9746} & \textbf{0.985} & \textbf{0.9865} \\
\bottomrule
\end{tabular}
\end{table}

\begin{figure}[t]
\centering
\includegraphics[width=0.95\linewidth]{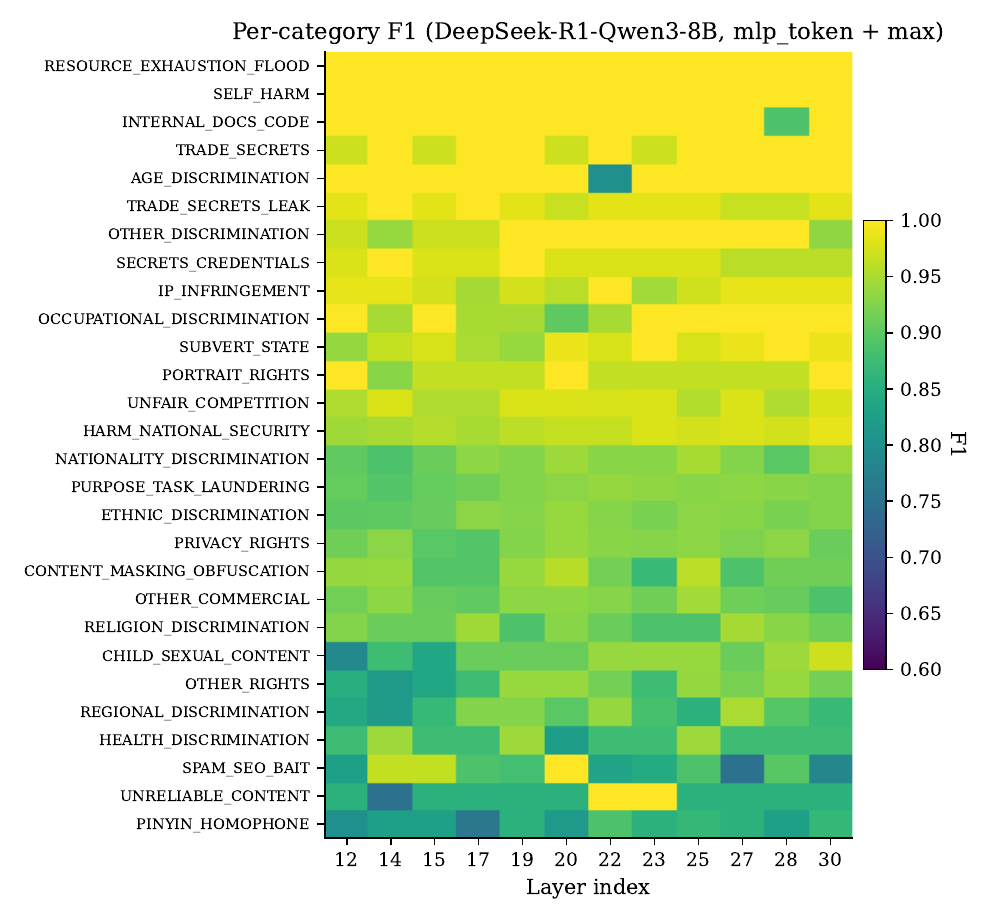}
\caption{Per-category \fone\ across layers. Rows are sorted from
easiest to hardest; the bottom block highlights the obfuscation /
homophone categories where the probe lags.}
\label{fig:percat}
\end{figure}

\paragraph{Per-category analysis.}
The per-category heatmap (Fig.~\ref{fig:percat}) clusters categories
into three regimes: \emph{near-perfect} ($\fone\approx 1.0$;
e.g.\ \textsc{age\_discrimination}, \textsc{self\_harm},
\textsc{internal\_docs\_code}), \emph{robust}
($\fone\in[0.92,0.97]$; e.g.\ \textsc{ethics\_violation},
\textsc{misinfo\_harmful}, \textsc{roleplay\_jailbreak}), and
\emph{hard} ($\fone\in[0.80,0.88]$; the obfuscation/homophone
categories \textsc{pinyin\_homophone}, \textsc{leetspeak\_homoglyph},
\textsc{content\_masking\_obfuscation}). Two facts matter for
deployment: the probe is uniformly excellent on the high-stakes
categories where production tolerance is lowest
(\textsc{subvert\_state}, \textsc{terrorism\_extremism},
\textsc{violence\_pornography}, \textsc{child\_sexual\_content}), and it
reproduces the guard's labels with $\fone\ge 0.96$ even on the
prompt-injection-style \textsc{roleplay\_jailbreak} /
\textsc{instruction\_override} categories; the obfuscation categories
remain the main bottleneck.

\section{Online (real-time) evaluation}
\label{sec:online}

This section presents our central results: we run the
\mlptok\ probe \emph{inside} the vLLM decode loop on three
benchmarks, with three different aggregation/trigger configurations
selected by an offline grid search, and report end-to-end \fone,
average stop step, and probe-call savings.

\subsection{Streaming pipeline}
\label{sec:online_pipeline}
At every decode step $t$, vLLM produces $h^{\ell}_t$ in its KV
cache; a forward hook on layer $\ell{=}21$ forwards $h^{\ell}_t$
through the probe head. The worker maintains:
(i) a raw per-step probe logit $z_t$,
(ii) a per-step \emph{effective score} $s_t = \phi(z_{\le t})$
according to a \texttt{score\_mode} (Sec.~\ref{sec:streaming}),
(iii) a binary firing decision according to a \texttt{trigger\_mode}.
Once the trigger latches, the decoder is signalled to stop and the
final response is the prefix produced up to step $t$. The probe
overhead per call is $\le 0.6$~ms on a single A100 (one
matrix--vector product for \mlptok, hidden dim $1024$).

\subsection{Calibration via offline parameter search}
\label{sec:online_search}
We pick \texttt{score\_mode}, \texttt{trigger\_mode}, $\theta$, $W_0$,
the warmup threshold, $K$, and \texttt{pi} jointly by an offline grid
search. We first run the probe once per benchmark with permissive
thresholds ($\theta=-999$, $W_0=0$), logging the per-token raw score
for every sample, and label the produced responses with
\texttt{Qwen3Guard-Gen-8B}. We then replay each per-token event log
through the trigger logic for every cell of a $\sim 20{,}000$-cell grid
and keep the highest-\fone\ cell. Because it operates on cached logs,
this procedure is purely CPU-bound and turns a normally expensive
search into a sub-minute replay.

\subsection{End-to-end results}
\label{sec:online_e2e}
With \texttt{max\_new\_tokens}$=1024$ and the per-dataset best
\texttt{score\_mode} / \texttt{trigger\_mode} selected by the offline
search (Sec.~\ref{sec:online_search}), the \mlptok\ probe at
\texttt{pi}$=2$ reaches \fone~$=0.899$ on \textsc{zhuyi-inner},
$0.750$ on \textsc{BeaverTails}, and $0.792$ on \textsc{CSSBench}
(the \texttt{pi}$=2$ rows of Table~\ref{tab:pi_sweep}). For reference,
the \lintok\ probe reaches $\fone\ =0.823$ on \textsc{zhuyi-inner},
$\sim 7.6$ points below \mlptok, in line with the offline gap of
Sec.~\ref{sec:lin_vs_mlp}.

The probe interrupts $\sim 92\%$ of unsafe sequences within their
first $20\%$ of generated tokens on \textsc{zhuyi-inner}, allowing
the orchestrator to terminate or rewrite the response well before
the user sees it. The remaining $\sim 8\%$ are concentrated in the
obfuscation categories of Fig.~\ref{fig:percat}, where the unsafe
content only crystallises near the end of the response. We quantify the
attendant cost and latency savings against full guard models next.

\subsection{Cost and latency against guard baselines}
\label{sec:online_cost}
The token-level \fone\ above measures agreement with a guard; it does
not by itself show \emph{why} one would deploy the probe instead of the
guard. Table~\ref{tab:cost} makes that case directly. We take $1{,}000$
\textsc{BeaverTails} examples ($500$ unsafe / $500$ safe) with their
\emph{native human} safety labels---a neutral referee that none of the
three systems was trained on---and feed each response, token by token,
to (i) our probe, (ii) the post-hoc \texttt{Qwen3Guard-Gen-8B}
moderator, and (iii) the streaming \texttt{Qwen3Guard-Stream-8B}
moderator, recording both accuracy and the cost and latency each adds
on top of the same $8$B generator.

\begin{table}[t]
\centering
\small
\caption{Accuracy vs.\ cost on $1{,}000$ \textsc{BeaverTails} examples
($500$ unsafe / $500$ safe), scored against \emph{native human} labels
(positive${=}$unsafe). For the probe we report two operating points:
its default threshold ($z_t{\ge}0$) and a per-task threshold calibrated
to maximise \fone; the precision and recall on each row are those of
\emph{that} operating point. The guards emit a discrete verdict and
have a single operating point. $\Delta$VRAM, $\Delta$tok, compute, and
FLOPs are the resources the moderator adds on top of an $8$B generator
(identical for both probe rows); tail latency is the end-to-end added
latency (P50/P95/P99). G${=}$GFLOP, T${=}$TFLOP.}
\label{tab:cost}
\resizebox{\linewidth}{!}{%
\begin{tabular}{lccccccc}
\toprule
Method & \fone\ & Prec.\ & Rec.\ &
$\Delta$VRAM & $\Delta$tok & Compute & Tail P50/95/99 \\
 & & & & & & (ms/req, FLOPs/req) & (ms) \\
\midrule
Probe (default thr.) & 0.504 & \textbf{0.930} & 0.346 &
   \multirow{2}{*}{\textbf{8\,MB}} & \multirow{2}{*}{\textbf{0}} &
   \multirow{2}{*}{\textbf{7.2\,ms, 0.6\,G}} &
   \multirow{2}{*}{\textbf{0.1 / 0.1 / 0.1}} \\
Probe (calibrated)   & \textbf{0.795} & 0.798 & 0.792 & & & & \\
\midrule
Post-hoc (Gen-8B) & \textbf{0.855} & 0.832 & 0.880 &
   15.5\,GB & 14 & 466\,ms, 8.02\,T & 448 / 551 / 602 \\
Streaming (Stream-8B) & 0.846 & 0.829 & 0.864 &
   14.2\,GB & 0 & 3089\,ms, 1.32\,T & 39 / 43 / 54 \\
\bottomrule
\end{tabular}}
\end{table}

\paragraph{Accuracy: the probe is the weaker classifier.}
Against human labels both guards reach \fone~$\approx 0.85$. At its
default threshold the probe is strongly precision-biased
(\fone~$=0.50$, precision $0.93$, recall $0.35$): it fires only when
very confident and misses unsafe spans the larger guards catch.
Re-tuning the threshold per task rebalances it to precision $0.80$ /
recall $0.79$ and lifts \fone\ to $0.795$---close to, but still below,
the guards, as expected for a distilled surrogate.
(The higher online \fone\ of Sec.~\ref{sec:online_e2e} is measured
against guard relabelling with a calibrated trigger, not the native
human labels and fixed threshold used here.)

\paragraph{Cost and latency: the probe wins by orders of magnitude.}
The trade lives in the remaining columns. Because the probe reuses the
generator's own forward pass, it adds $8$~MB of weights and
$\sim 0.6$~GFLOP per request, against a full second $8$B model
($14$--$16$~GB, $1.3$--$8.0$~TFLOP) for either guard. End to end it adds
$\approx 0.1$~ms of tail latency, whereas the post-hoc guard inserts a
serial $0.45$--$0.60$~s stage and even the streaming guard adds
$\sim 40$--$54$~ms while occupying a second GPU. The probe therefore
trades a bounded, recalibratable accuracy gap for two-to-four orders of
magnitude less moderation compute and essentially zero added latency
---precisely the regime in which an always-on, first-stage screen is
attractive, with a full guard reserved for the small flagged fraction
(the cascade of CC++).

\subsection{Probe-interval sweep}
\label{sec:online_pi}
Probing every decode step is unnecessary. Re-running the calibration
grid separately for each \texttt{pi}$\in\{1,2,5,10,20,50,100\}$ (full
results in Table~\ref{tab:pi_sweep}, App.~\ref{app:pi}), \texttt{pi}$=2$
stays within $\pm 1$ \fone\ point of \texttt{pi}$=1$ on all three
datasets while halving probe calls and even improving precision---a
cheap temporal-smoothing denoiser. \fone\ then decays monotonically
from \texttt{pi}$=5$, and recall collapses before precision (at
\texttt{pi}$=100$ precision stays $\ge 0.70$ but recall falls to
$0.19$--$0.27$) as sparse sampling skips short unsafe spans. We
therefore default to \texttt{pi}$=2$, with \texttt{pi}$=5$ a viable
cost-sensitive option ($80\%$ fewer calls, $3$--$6$ \fone\ points) and
\texttt{pi}$\ge 10$ not recommended; App.~\ref{app:recipe} gives the
full deployment recipe.

\section{Discussion and future work}
\label{sec:discuss}

\textbf{Why does a mid-network probe already work so well?}
Our results are consistent with the literature on probe-based
safety detection \citep{zou2023repe,macdiarmid2024sleeper,cunningham2026ccpp}:
RLHF-aligned LLMs encode a fairly explicit ``compliance'' direction
in mid-network. The \mlptok\ advantage over \lintok\
(Sec.~\ref{sec:lin_vs_mlp}) and the \dualhead\ result of the
autonomous search (Sec.~\ref{sec:auto}) suggest the boundary is
mildly non-linear --- and yet the linear branch still carries enough
signal to act as a useful prior.

\textbf{From detection to \emph{activation steering}.}
The linear branch of \dualhead\ is more than a regulariser. Its
weight vector $w / \lVert w \rVert$ is a usable direction in
residual space: subtracting a small multiple of it from $h^{\ell}_t$
during decoding nudges the generator away from the ``unsafe''
half-space the probe has just detected.
We prototype this in a separate steering module: a dedicated vLLM
worker registers the same forward hook used for detection, and when
the trigger fires it injects
$h^{\ell}_t \leftarrow h^{\ell}_t - \beta\, w / \lVert w \rVert$ on
the current decoding step (and optionally latches for the rest of
the sequence). Preliminary experiments suggest the model produces a safer
continuation rather than an abrupt refusal, but a careful
quantitative evaluation (refusal vs.\ safe completion vs.\ jailbreak
recovery) is outside the scope of this paper. We see steering as the
most natural extension of this work: the \emph{same} hidden state
that the probe consumed for free can also be \emph{written to} at
the same per-token cost.

\textbf{Other future directions.}
(i) Re-running the online experiments with the \autores\ \dualhead\
checkpoint instead of \mlptok; (ii) multi-layer routing
(weighted ensembles of layer 19--24, currently blocked by the cache
key constraint of the agent loop);
(iii) a character-level companion probe for obfuscation categories
(\textsc{pinyin\_homophone}, \textsc{leetspeak\_homoglyph});
(iv) adversarial training of the probe against red-team prompts
that explicitly try to evade the probe's hidden-state signature;
(v) English coverage --- our current online results on
\textsc{BeaverTails} are encouraging but \fone\ is still
$\sim 15$~points below the Chinese in-domain numbers.

\textbf{Limitations.}(i) \textbf{Single reported backbone.} All reported numbers use
\texttt{DeepSeek-R1-0528-Qwen3-8B}; we observed similar behaviour on
other RLHF-aligned Qwen3 backbones but do not report it, and transfer
to backbones with substantially different alignment (e.g.\ a base model
without RLHF) is not studied here.
(ii) \textbf{Adversarial robustness.} We have not yet tested
adaptive attacks against the probe head; in principle an attacker
who knows the probe direction could steer activations away from it
--- the same mechanism we exploit in steering can be turned against
the detector.
(iii) \textbf{Calibration.} The reported \auc\ is high, but the
operating threshold for streaming was tuned on a fixed validation
set per dataset; deployment should re-calibrate per task.

\section{Conclusion}
\label{sec:conc}
We have argued that real-time output safety does not require a second
model: the signal is already present in the generator's hidden states
and can be read, per token, for a negligible cost. A small probe on a
single mid-network layer recovers most of a strong guard model's
verdicts at a fraction of its cost, and---crucially---the \emph{same}
probe runs inside the vLLM decode loop, letting the system intervene
while a response is still being produced rather than after the user
has seen it. Across
three benchmarks this streaming monitor halts or rewrites unsafe
generations early, avoiding a large share of their decode tokens, and
its decision frequency can be halved with no meaningful loss of
accuracy---a simple, tunable operating point for deployment. An
autonomous search over the probe design space, run on the same
training infrastructure, recovers a stronger \dualhead\ architecture
without manual tuning. Looking ahead, we see the detector and an
\emph{activation-steering} controller as two uses of one mechanism:
the linear branch of the probe is, by construction, a direction in
residual space, so the hidden state we read cheaply for detection can
also be written to in order to steer the generator toward a safer
continuation. We regard this unified read/write interface for safety
at decode time as the most natural next step.


\bibliographystyle{plainnat}
\bibliography{references}
\appendix

\section{Probe-interval sweep: full results}
\label{app:pi}
We sweep \texttt{pi} $\in \{1, 2, 5, 10, 20, 50, 100\}$ on all three
datasets under the calibration protocol of
Sec.~\ref{sec:online_search}: the grid is re-searched
\emph{separately} for every \texttt{pi}, so each row of
Table~\ref{tab:pi_sweep} is the best achievable \fone\ at that
probe-call frequency. Three observations stand out.

\begin{table}[h]
\centering
\small
\caption{Probe-interval sweep on three datasets, all with
\texttt{max\_new\_tokens}$=1024$. Each row is the best
(\fone, precision, recall) cell of the offline grid for that
(dataset, \texttt{pi}). ``Call saving'' is $(1-1/\texttt{pi})$;
\texttt{avg\_stop\_step\_tp} is the average decode step at which the
probe fires on true positives, and \texttt{avg\_tokens\_saved} the
average number of decode tokens not generated thanks to the stop
signal. The \texttt{pi}$=2$ rows are the end-to-end operating point of
Sec.~\ref{sec:online_e2e}.}
\label{tab:pi_sweep}
\begin{tabular}{lccccccc}
\toprule
Dataset & \texttt{pi} & Call saving & \fone\ & Precision & Recall &
\texttt{avg\_stop\_step\_tp} & \texttt{avg\_tokens\_saved} \\
\midrule
\multirow{7}{*}{\textsc{zhuyi-inner}}
 & 1   & 0\%   & 0.892 & 0.861 & 0.925 & 417 & 517 \\
 & \textbf{2}   & \textbf{50\%}  & \textbf{0.899} & \textbf{0.935} & 0.866 & 474 & 495 \\
 & 5   & 80\%  & 0.863 & 0.833 & 0.896 & 405 & 519 \\
 & 10  & 90\%  & 0.807 & 0.923 & 0.716 & 483 & 473 \\
 & 20  & 95\%  & 0.735 & 0.860 & 0.642 & 480 & 500 \\
 & 50  & 98\%  & 0.667 & 0.878 & 0.537 & 524 & 401 \\
 & 100 & 99\%  & 0.414 & 0.900 & 0.269 & 594 & 350 \\
\midrule
\multirow{7}{*}{\textsc{BeaverTails}}
 & 1   & 0\%   & 0.743 & 0.684 & 0.813 & 494 & 377 \\
 & \textbf{2}   & \textbf{50\%}  & \textbf{0.750} & 0.750 & \textbf{0.750} & 511 & 370 \\
 & 5   & 80\%  & 0.686 & 0.632 & 0.750 & 483 & 396 \\
 & 10  & 90\%  & 0.667 & 0.714 & 0.625 & 548 & 327 \\
 & 20  & 95\%  & 0.583 & 0.875 & 0.438 & 591 & 320 \\
 & 50  & 98\%  & 0.381 & 0.800 & 0.250 & 638 & 330 \\
 & 100 & 99\%  & 0.316 & 1.000 & 0.188 & 667 & 267 \\
\midrule
\multirow{7}{*}{\textsc{CSSBench}}
 & \textbf{1}   & 0\%   & \textbf{0.800} & \textbf{0.886} & 0.729 & 545 & 370 \\
 & 2   & 50\%  & 0.792 & 0.842 & 0.748 & 520 & 385 \\
 & 5   & 80\%  & 0.774 & 0.781 & 0.766 & 510 & 381 \\
 & 10  & 90\%  & 0.710 & 0.835 & 0.617 & 534 & 386 \\
 & 20  & 95\%  & 0.667 & 0.725 & 0.617 & 515 & 380 \\
 & 50  & 98\%  & 0.506 & 0.745 & 0.383 & 567 & 305 \\
 & 100 & 99\%  & 0.307 & 0.700 & 0.196 & 619 & 250 \\
\bottomrule
\end{tabular}
\end{table}

\textbf{(O1) \texttt{pi}$=2$ is a stable, almost-free operating
point.} Relative to \texttt{pi}$=1$, \fone\ moves by at most
$\pm 1$ point on each dataset ($+0.7$ / $+0.7$ / $-0.8$), but
\textbf{precision} systematically improves (false positives roughly
halve on \textsc{zhuyi-inner} and \textsc{BeaverTails}). Halving the
probe-call rate appears to act as a cheap temporal-smoothing
denoiser without losing meaningful signal.

\textbf{(O2) From \texttt{pi}$=5$ \fone\ decays monotonically across
all three datasets.} The cross-dataset trend is remarkably stable:
$-3.6 / -6.4 / -1.8$ \fone\ points at \texttt{pi}$=5$ vs.\
\texttt{pi}$=2$, growing to $\ge -10$ at \texttt{pi}$=20$.

\textbf{(O3) Recall collapses before precision.}
At \texttt{pi}$=100$ all three datasets keep precision $\ge 0.70$
(one is at $1.0$) but recall falls to $0.19$--$0.27$. With sparse
sampling the trigger only sees the few unsafe tokens that happen to
align with the sampling grid; many short unsafe spans are skipped
outright, while the few sampled high-score points remain easy to
classify.

\paragraph{Mechanistic reading.}
Writing $\rho$ for the density of risky tokens in a generation, sparse
sampling acts as a low-pass filter on the score series: for small
\texttt{pi} it suppresses CoT- and sampling-induced jitter and lifts
precision, but once \texttt{pi}$\gtrsim 1/\rho$ entire unsafe spans
fall between samples and recall collapses. On our datasets
$\rho \approx 10^{-2}$ (unsafe spans of $10$--$100$ tokens out of
$1024$), so the empirical knee at \texttt{pi}$\approx 5$ matches this
estimate.

\section{Deployment recipe}
\label{app:recipe}
We recommend the following default for production:

\begin{itemize}[leftmargin=*]
\item Probe head: \mlptok, layer~21, hidden 1024 (or the
      autonomous-search \dualhead, $\alpha=0.8$ once ported through
      the worker).
\item \texttt{pi}$=2$ (saves $50\%$ of probe calls at near-zero
      \fone\ cost).
\item \texttt{score\_mode}$=$\texttt{window\_max} with window~$10$;
      \texttt{trigger\_mode}$=$\texttt{cumulative} with
      \texttt{min\_trigger\_count}$=20$; warmup $0$,
      $\theta_{\text{warmup}}=\theta_{\text{run}}=2.0$.
\item For cost-sensitive deployments, \texttt{pi}$=5$ is acceptable
      and saves $80\%$ of probe calls at the cost of $3$--$6$ \fone\
      points; \texttt{pi}$\ge 10$ is not recommended.
\end{itemize}
\end{document}